\PassOptionsToPackage{table}{xcolor}

\documentclass{article}
\usepackage[utf8]{inputenc}
\usepackage{microtype}
\usepackage{graphicx}
\usepackage{booktabs} 
\usepackage{comment}
\usepackage{arydshln}
\usepackage{makecell}
\usepackage{hyperref}
\usepackage{mathtools}
\usepackage{color}
\usepackage{listings}
\usepackage{xcolor}

\usepackage{float}
\newfloat{listing}{tbhp}{lol}
\floatname{listing}{Listing}

\usepackage{graphicx}
\usepackage{CJKutf8}
\usepackage{url}
\usepackage{multicol,multirow}
\usepackage{makecell}
\usepackage{algorithmic}
\usepackage{algorithm}
\usepackage{booktabs}
\usepackage{textcomp}

\usepackage{bm}
\usepackage{parskip}
\usepackage[T1]{fontenc}    
\usepackage{booktabs}       
\usepackage{amsmath, amsthm}
\usepackage{amssymb}
\usepackage{amsfonts}
\usepackage{times}
\usepackage{latexsym}
\usepackage{amsmath, amsthm}
\usepackage{amssymb}
\usepackage{amsfonts}

\usepackage{tabularx}

\usepackage{tabularx}       
\usepackage{verbatim}       
\usepackage{fancyvrb}       
\usepackage{listings}       
\usepackage{enumitem}       
\usepackage{amsmath}        
\usepackage{geometry}       

\usepackage{booktabs}       
\usepackage{float}          
\usepackage{caption}        

\usepackage[most]{tcolorbox}

\usepackage{longtable}
\usepackage{listings}
\usepackage{booktabs, listings, enumitem}
\usepackage{listings}
\usepackage{graphicx}
\usepackage{adjustbox}

\definecolor{promptblue}{RGB}{200, 230, 255}
\definecolor{exemplargreen}{RGB}{220, 255, 220}
\definecolor{protocolyellow}{RGB}{255, 255, 200}

\setlist{nolistsep}

\usepackage{booktabs,  tcolorbox, listings, etoolbox, tabularx}
\tcbuselibrary{listings, skins, breakable}

\usepackage{fontawesome5}

\definecolor{taskblue}{RGB}{0,99,177}
\definecolor{refgreen}{RGB}{0,150,85}
\definecolor{subviolet}{RGB}{131,76,190}
\definecolor{templateblue}{RGB}{1, 128, 134}
\definecolor{refgreenDark}{RGB}{0, 90, 45}
\definecolor{analysisblue}{HTML}{1E90FF} 
\definecolor{algoyellow}{HTML}{A0522D}   

\lstdefinestyle{codestyle}{
  language=Python,
  basicstyle=\footnotesize\ttfamily,
  frame=single,
  numbers=left,
  numberstyle=\tiny,
  xleftmargin=1.5em,
  framexleftmargin=1.5em,
  keywordstyle=\color{taskblue},
  commentstyle=\itshape\color{gray},
  stringstyle=\color{orange},
  showstringspaces=false,
  breaklines=true,
  tabsize=2
}
\definecolor{kwblue}{HTML}{005CFF}       
\definecolor{strred}{HTML}{B80034}    
\definecolor{codebg}{RGB}{245,248,250}
\definecolor{argsc}{RGB}{0,128,128}
\definecolor{codegreen}{HTML}{189399}

\lstdefinestyle{py}{
  language        = Python,
  basicstyle      = \tiny\ttfamily,
  numbers         = left,
  numberstyle     = \tiny\color{gray},
  stepnumber      = 1,
  keywordstyle    = \color{kwblue}\bfseries,
  commentstyle    = \color{gray},
  stringstyle     = \color{strred},
  breaklines      = true,
  showstringspaces= false,
  tabsize         = 4,
  backgroundcolor = \color{templateblue!5},
  frame           = single,
  xleftmargin     = 3em,             
  framexleftmargin= 2.5em,  
  rulecolor       = \color{codebg},
  framesep        = 6pt,        
  literate        = {Args}{{\textcolor{argsc}{Args}}}4
}
\usepackage{longtable,booktabs} 
\colorlet{subvioletstrong}{subviolet!80!black} 
\usepackage{listings}
\lstdefinestyle{afteroptimcode}{
  language=Python, numbers=left, frame=none,
  numberstyle   = \tiny,        
  numbersep     = 3pt,          
  xleftmargin   = 2pt,          
  framexleftmargin = 0pt,       
  basicstyle=\ttfamily\footnotesize, keywordstyle=\color{subviolet},
  breaklines=true, columns=fullflexible
}

\usepackage{listings}
\lstdefinestyle{beforeoptimcode}{
  language=Python, numbers=left, frame=none,
  numberstyle   = \tiny,        
  numbersep     = 3pt,          
  xleftmargin   = 2pt,          
  framexleftmargin = 0pt,       
  basicstyle=\ttfamily\footnotesize, keywordstyle=\color{subviolet},
  breaklines=true, columns=fullflexible
}

\usepackage[normalem]{ulem} 
\usepackage{multirow}
\usepackage{nicematrix} 
\usepackage{pgfplots}   
\usepackage{xcolor}     
\usepackage{array}      
\usepackage{arydshln}
\definecolor{code-highlight-blue}{HTML}{2696f0}
\definecolor{code-highlight-green}{HTML}{7eb547}
\definecolor{code-highlight-yellow}{HTML}{fdcc3b}
\definecolor{code-highlight-purple}{HTML}{ab4abb}
\definecolor{code-highlight-red}{HTML}{f3473a}
\definecolor{code-highlight-orange}{HTML}{fe970c}

\DeclareCaptionLabelFormat{figsub}{Figure~\thefigure(\alph{subfigure})}

\newcommand{\emailmark}{%
    \textsuperscript{\large\Letter}%
}

\usepackage{pgfplots}
\usepackage{subcaption}

\usepackage{marvosym}

\newcommand{\emailtext}[1]{%
    \begingroup
    \renewcommand{\thefootnote}{\large\Letter}%
    \footnotetext{#1}%
    \endgroup
}

\usepackage{wrapfig}

\title{CUDA-L2: Surpassing cuBLAS Performance for Matrix Multiplication through Reinforcement Learning}

\author{Songqiao Su, Xiaoya Li, Albert Wang, Guoyin Wang, Jiwei Li and Chris Shum}

\date{\textbf{\large Ornith Team}\\\vspace{0.1cm}\includegraphics[scale=0.05]{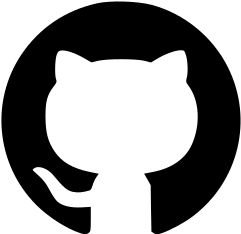} 
\href{https://github.com/ornith-ai/CUDA-L2}{{\large github.com/ornith-ai/CUDA-L2}}}

\usepackage{listings}
\usepackage{xcolor}

\usepackage{booktabs}
\usepackage[table]{xcolor}

\definecolor{rowcream}{HTML}{FFFACD}
\definecolor{rowyellow}{HTML}{FFF8DC}
\definecolor{headercolor}{HTML}{E0E0E0}

\begin{document}

\maketitle

\begin{abstract}

Matrix multiplication (matmul) is one of the most fundamental operations in LLMs. However, manually optimizing  Matmul kernels is  challenging due to the fact that different matrix dimension (M, N, K) require different optimization strategies and that optimizations rarely transform across different GPU architectures, which make comprehensive manual tuning hard at scale. 
In this paper, we propose CUDA-L2, a system that combines large language models (LLMs) and reinforcement learning (RL) to automatically optimize Half-precision General Matrix Multiply (HGEMM) CUDA kernels. 
Using CUDA execution speed as the RL reward, CUDA-L2 automatically optimizes HGEMM kernels across 1,000 configurations. These configurations represent all $10^3$ combinations of M, N, K values from \{64, 128, 256, 512, 1024, 2048, 4096, 8192, 12288, 16384\}, and already covers those 
used in attention and FFN layers of widely open-sourced  models like Qwen, Llama and DeepSeek.

CUDA-L2 systematically outperforms major matmul baselines to date,  from the widely-used {\it torch.matmul} to state-of-the-art Nvidia's closed-source libraries, i.e., {\it cuBLAS}, {\it cuBLASLt}. In  offline mode, where kernels are executed consecutively without time intervals, CUDA-L2 yields +22.0\% over {\it torch.matmul} on average; +19.2\% over 
 {\it cuBLAS} using the optimal layout configuration 
(normal-normal NN and transposed-normal TN); +16.8\% over {\it cuBLASLt-heuristic}, which queries  {\it cuBLASLt} library and selects the algorithm based on the heuristic's suggestion; and +11.4\% over the most competitive {\it cuBLASLt-AutoTuning} model, which selects the fastest algorithm from up to 100 candidates from {\it cuBLASLt}'s suggestions. In  server mode, where kernels are executed at random intervals simulating real-time inference, the speedups further increase  to +28.7\%, +26.0\%, +22.4\%, and +15.9\% for {\it torch.matmul}, {\it cuBLAS}, {\it cuBLASLt-heuristic}, and {\it cuBLASLt-AutoTuning} respectively.

CUDA-L2 shows that even the most performance-critical, heavily-optimized kernels like HGEMM can be improved through LLM-guided RL automation by systematically exploring configuration spaces at scales impractical for humans.
 While the current version of CUDA-L2 only focuses on A100 GPUs, the framework is designed for broad applicability, with ongoing work to extend it to other GPU architectures, including  Ada Lovelace, Hopper and Blackwell.\emailmark
\end{abstract}
\begin{figure*}[!h]
 \centering
 \begin{adjustbox}{margin=-0.5cm 0cm 0cm 0cm}
 \begin{minipage}[c]{\textwidth}
 \centering
\includegraphics[scale=0.4]{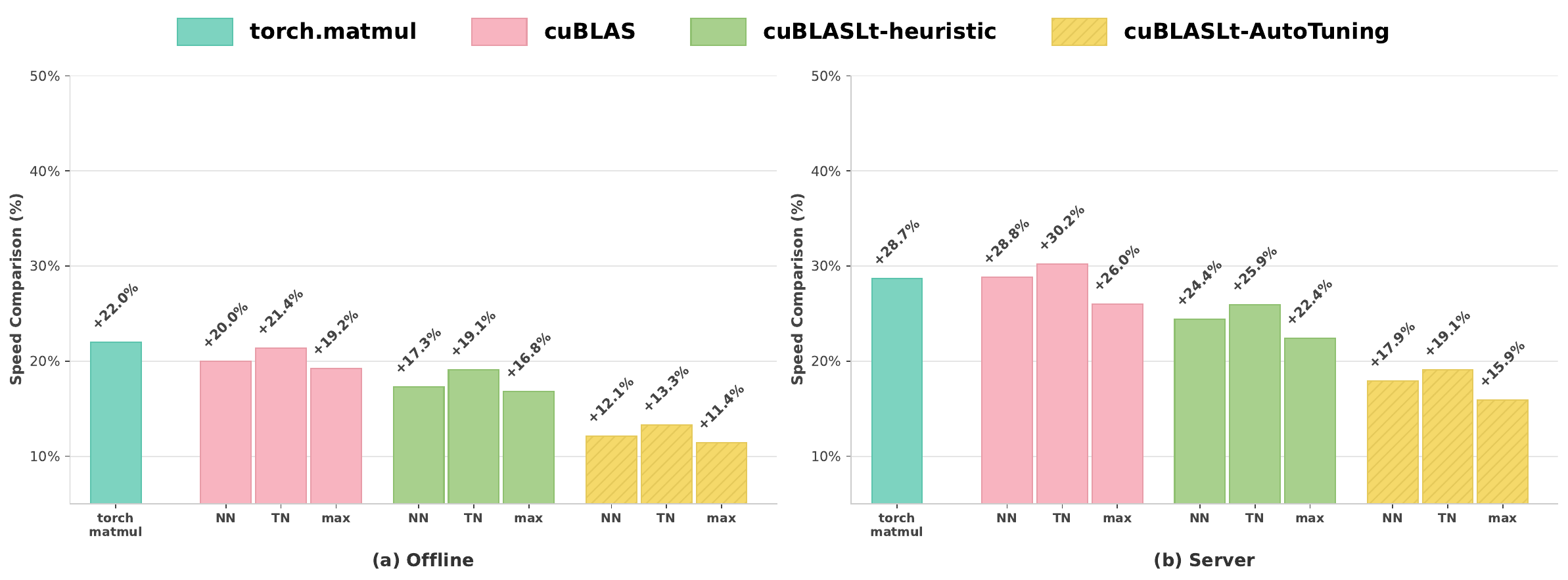}
\caption{
Benchmark results on 1,000 (M, N, K) configurations for Half-precision General Matrix Multiply (HGEMM). We report CUDA-L2's performance against
{\it torch.matmul}, 
 {\it cuBLAS}, {\it cuBLASLt-heuristic}, and {\it cuBLASLt-AutoTuning} for NN, TN, and max(NN,TN) layouts: (a) Offline scenario: kernel executed consecutively without time intervals. (b) Server scenario: kernel executed at  intervals, simulating real-time inference.}
\label{fig:average_speedup}
 \end{minipage}
 \end{adjustbox}
 \end{figure*}

\emailtext{~Email: \{songqiao\_su, xiaoya\_li, albert\_wang, jiwei\_li, chris\_shum\}@ornith.ai}
\section{Introduction}
Matrix multiplication (matmul) is one of the most fundamental operations in deep learning and LLMs and accounts for a very significant portion of computation time in both training and inference. 
Consequently, its CUDA kernels are among the most deeply optimized by human experts. 

However, manually optimizing matmul kernels across diverse matrix sizes remains challenging due to the vast configuration space and architecture-specific constraints: different matrix dimension triplets (M, N, K) may require distinct optimization strategies, and these strategies rarely transfer across GPU architectures due to different hardware characteristics.
Moreover, even for identical matrix dimensions on the same GPU, optimal kernel can be different depending on 
setups such as
accumulator precision, where using FP16 versus FP32 accumulators (both valid choices for FP16 inputs) leads to different register pressure and thus different optimization strategies.
Therefore, despite being deeply optimized, matmul is far from a solved problem, and significant performance gains are still possible.
This is evidenced by the 13\% speedup from targeted Grouped GEMM optimizations for the deepseek-R1 model by the Tensor-RT LLM team\footnote{\url{https://nvidia.github.io/TensorRT-LLM/blogs/tech_blog/blog3_Optimizing_DeepSeek_R1_Throughput_on_NVIDIA_Blackwell_GPUs.html}}. This speedup was achieved when human expertise was narrowly focused on a single target. 
Systematically improving matmul kernels calls for automated optimization methods that can scale across the vast space of problems and GPU architectures.

Rapid advancements in LLMs \cite{achiam2023gpt,comanici2025gemini,guo2025deepseek, team2023gemini,yang2025qwen3} open the door to automatic kernel generation, and
the past year has witnessed a growing interest in harnessing LLMs, especially RL-augmented LLMs, for autonomous CUDA kernel generation \cite{ouyang2025kernelbench, baronio2025kevin, li2025cuda, lange2025ai, tschand2025swizzleperf}. Existing work, such as Sakana AI's AI CUDA Engineer \cite{lange2025ai} and CUDA-L1 \cite{li2025cuda}, primarily optimize kernels from benchmarks like KernelBench \cite{ouyang2025kernelbench}, which covers a diverse range of CUDA operations, each evaluated on a single, fixed configuration (e.g., one specific input dimension). However, it remains unclear how these benchmark-optimized kernels translate to real-world production environments. To the best of our knowledge, no prior work has achieved performance comparable to manually-optimized matmul kernels, particularly when compared against NVIDIA's highly-optimized {\it cuBLAS} library.

In this work, we propose CUDA-L2, a system that combines LLMs and RL to automatically optimize matmul CUDA kernels. 
CUDA-L2 
covers 1,000 configurations that represent all $10^3$ combinations of M, N, K values from \{64, 128, 256, 512, 1024, 2048, 4096, 8192, 12288, 16384\}, 
the size of
which already covers those 
used in attention and FFN layers of  open-sourced  models like Qwen \cite{yang2025qwen3},
Llama \cite{dubey2024llama}, 
 and DeepSeek \cite{guo2025deepseek}. 
To tackle the challenging task of matrix multiplication, 
CUDA-L2 extends CUDA-L1 through several technical enhancements, including (1) continued pretraining with more diverse CUDA code; (2) multi-stage RL training progressing from broad (general kernel types) to specialized domains (matmul-specific kernels);  
(3) including more comprehensive NCU (NVIDIA Nsight Compute) profiling metrics (e.g., memory throughput, SM occupancy, cache efficiency) in the context for better optimization decisions; and (4)
incorporating
retrieval-augmented context to accommodate new knowledge or architectural characteristics not covered in the foundation model.

CUDA-L2 systematically outperforms major matmul baselines to date, ranging from the widely-used {\it torch.matmul} to state-of-the-art optimized libraries ({\it cuBLAS}, {\it cuBLASLt}). We evaluate under two scenarios: (1) offline, where kernels are executed consecutively without time intervals, and (2) server, where kernels are executed at random intervals simulating real-time inference. In the offline scenario, across 1,000 configurations, CUDA-L2 yields +22.0\% over {\it torch.matmul} on average; +19.2\% over {\it cuBLAS} using its optimal layout configuration (NN or TN); +16.8\% over {\it cuBLASLt-heuristic}, which queries the {\it cuBLASLt} library and selects the algorithm based on the heuristic's suggestions; and +11.4\% over the most competitive {\it cuBLASLt-AutoTuning} setup, which selects the top-performing algorithm from up to 100 candidates from {\it cuBLASLt}'s heuristic API. In the server scenario, the speedups further increase to +28.7\%, +26.0\%, +22.4\%, and +15.9\% for {\it torch.matmul}, {\it cuBLAS}, {\it cuBLASLt-heuristic}, and {\it cuBLASLt-AutoTuning} respectively.

\section{Preliminaries}  
In this section, we first describe preliminaries for developing HGEMM kernels. 
\subsection{HGEMM}
\paragraph{HGEMM}
 Half-precision General Matrix Multiplication (HGEMM) is  one of the most widely used matmul kernels in current LLMs. It
  computes the matrix product $C = \alpha AB + \beta C$ using 16-bit  arithmetic.
An $(M, N, K)$ triplet denotes a dimension configuration for HGEMM, where matrix ${A} \in \mathbb{R}^{M \times K}$ and matrix ${B} \in \mathbb{R}^{K \times N}$.
In practice, the common case is $\alpha = 1$ and $\beta = 0$, yielding $C = AB$, which we focus on in this work. 
We adopt this setup throughout the paper.
An $(M, N, K)$ triplet denotes a dimension configuration where ${A} \in \mathbb{R}^{M \times K}$ and ${B} \in \mathbb{R}^{K \times N}$.

Modern HGEMM implementations follow a tiled approach to maximize data reuse. The computation is organized into three main phases:
First, the matrices A and B are divided into smaller tiles. Each thread block computes one tile of the output matrix C, with dimensions BM × BN. Within each thread block, tiles are further subdivided to match the GPU's tensor core dimensions.
Second, during the mainloop, data flows through multiple memory levels. Tiles of size BM × BK from A and BK × BN from B are loaded from global memory into shared memory, then from shared memory into registers. Once in registers, tensor cores perform the actual matrix multiply-accumulate operations. This process repeats across all tiles along the K dimension, with partial results accumulated in registers.
Finally, in the epilogue phase, the accumulated results are written back from registers to shared memory, and then from shared memory to global memory.
\subsection{Baselines for Comparison}
\subsubsection{torch.matmul}
PyTorch's {\it torch.matmul} naturally constitutes a baseline. PyTorch's implementation automatically selects optimized kernels based on input dimensions and data types. For half-precision operations, it dispatches to {\it cuBLAS} internally but includes additional overhead from PyTorch's tensor dispatch system and memory management. 
{\it torch.matmul}
represents a standard and robust practice for the majority of users who rely on default PyTorch operations. 
 
 \subsubsection{cuBLAS}
 NVIDIA’s {\it cuBLAS} library provides a strong optimized baseline and delivers high performance without requiring any manual tuning. We evaluate two
most widely-used
 matrix layouts: 
{\it cuBLAS-NN} (all row-major) and {\it cuBLAS-TN} (mixed layout).
For each (M, N, K) configuration, we also compare CUDA-L2 with {\it cuBLAS} using the optimal layout with respect to each dimension, i.e., max({\it cuBLAS-NN}, {\it cuBLAS-TN}), denoted by {\it cuBLAS-max}.
We use the  {\it cublasGemmEx} function offered by 
NVIDIA’s {\it cuBLAS} library
with the {\it CUBLAS\_GEMM\_DEFAULT\_TENSOR\_OP} operation to enable Ampere FP16 Tensor Cores and allow {\it cuBLAS}’s internal heuristics to automatically select the optimal algorithm. Code snippets for the 
{\it cuBLAS-NN}
 and {\it cuBLAS-TN} baselines are provided in Listing~\ref{lst:cublas_baseline} in Appendix.

  \subsubsection{cuBLASLt}
  {\it cuBLASLt} provides a lower-level, more controllable interface to NVIDIA’s optimized GEMM kernels than the higher-level cuBLAS. Unlike {\it cuBLAS}'s black-box heuristics that hide algorithm selection, {\it cuBLASLt} exposes all available algorithm variants to developers and allows for explicitly enumerating and evaluating them. Though this enables  tuning that can yield additional speedups over {\it cuBLAS}’s defaults,  it also demands substantial expertise and tuning effort.
We compare   CUDA-L2 with two 
  {\it cuBLASLt} setups based on API offered by Nvidia, {\it cuBLASLt-heuristic} and {\it cuBLASLt-AutoTuning}.
  
  \paragraph{cuBLASLt-heuristic} 
  The heuristic strategy
 uses NVIDIA's {\it cublasLtMatmulAlgoGetHeuristic} API, which is the standard recommended approach for {\it cuBLASLt} optimization. For each matrix configuration, it queries the library for algorithm recommendations and selects the top-ranked algorithm from the heuristic's suggestions, which is simply the recommendation with index 0. 
Similar to {\it cuBLAS},
for each dimension , 
we benchmark 
CUDA-L2 against {\it cuBLASLt-heuristic-NN}, {\it cuBLASLt-heuristic-TN} 
and  {\it cuBLASLt-heuristic-max}. 
To avoid 
the overhead introduced by
calling the API repeatedly during evaluation, the algorithm selection
for  NN and TN
 is performed beforehand. 
 The code for {\it cuBLASLt-heuristic-TN} is shown in Listing
\ref{lst:cublaslt_heuristic_tn}. 

  \paragraph{cuBLASLt-AutoTuning}
  AutoTuning\footnote{\url{https://github.com/NVIDIA/CUDALibrarySamples/tree/master/cuBLASLt/LtSgemmSimpleAutoTuning}}
employs an exhaustive strategy that first retrieves up to 100 algorithm candidates for each (M, N, K)  from {\it cuBLASLt}'s heuristic API {\it cublasLtMatmulAlgoGetHeuristic}.
Next, 
each returned algorithm is evaluated  with randomized execution order and random input matrixes. 
 The fastest algorithm is selected based on median execution time, as suggested by {\it cuBLASLt}'s documents. 
 Again, both NN and TN layouts are tested, using whichever achieves a faster speed. 
 Selected algorithms are cached to eliminate overhead during evaluation.

\subsection{Kernel Successfulness}

 A custom HGEMM kernel is successful if it is both executable and correct.
 \subsubsection{Executability} 
A kernel is executable if it successfully compiles, launches, and executes to completion within a reasonable time limit. 
{\it compute-sanitizer --tool memcheck} is used to  check for memory access violations.

\subsubsection{Correctness} 
To valid kernel correctness, we need to compare its output to a reference correct kernel with the same inputs. 
For HGEMM, we use the FP32 CPU implementation as the reference kernel due to its well-established correctness and numerical stability:
\begin{equation}
\text{ref} = \mathrm{MatMul}(\mathrm{float}(\mathrm{a.cpu()}),\, \mathrm{float}(\mathrm{b.cpu()})
\end{equation}
Theoretically, a kernel is correct if it produces outputs identical to those of a reference implementation. However, exact equivalence is not 
impossible
 on  GPUs due to the non-associative nature of floating-point arithmetic, i.e., $(a + b) + c \neq a + (b + c)$. To address this, we adopt 
 the following two practical criterion: 
\paragraph{Exact Match with binary Inputs.}
We randomly generate matrices $A, B$ with elements being binary in $\{0, 1\}$. For matrix multiplication $C = A\times B$, each output element is:
\[
c_{ij} = \sum_{k=1}^{K} a_{ik} \cdot b_{kj}
\]
Since $a_{ik}, b_{kj} \in \{0, 1\}$, each product is either 0 or 1, and the sum is guaranteed to be a non-negative integer. We first compute the reference output $C^{\text{ref}}$ using FP32 on CPU, which provides exact integer results. We then compute the output $C^{\text{custom}}$ using the custom kernel. For each position $(i,j)$ where $c^{\text{ref}}_{ij} < 2048$, we require $c^{\text{test}}_{ij} = c^{\text{ref}}_{ij}$ exactly,
and we ignore positions  $c^{\text{ref}}_{ij} > 2048$.
This is because
half-precision has 10 mantissa bits plus 1 implicit leading bit, yielding 11 bits of significand precision. Thus all integers in $[0, 2048)$ are exactly representable in half-precision, while integers $\geq 2048$ may not be.
Crucially, since each term $a_{ik} \cdot b_{kj} \in \{0, 1\}$, the partial sums are monotonically non-decreasing. If the final element value is below 2048, all intermediate values must also be below 2048, ensuring exactness for all intermediate steps when doing the addition for  $\sum_k a_{ik} \cdot b_{kj}$.\footnote{As a side note, sampling elements from $\{-1, 0, 1\}$ does not guarantee this property, since the partial sums are no longer monotonically non-decreasing. There can be cases where an intermediate sum exceeds 2048 (exactness already lost) before later cancellations bring the final result below 2048.}
\footnote{One might consider using fractional values such as $1/2$, $1/4$, or $1/8$, but this does not fundamentally change the analysis as it merely shifts the threshold. For example,
in FP16, the exactness threshold depends on the smallest representable increment at a given magnitude. 
Take 1/2 as an example, if elements are sampled from $\{0, 1/2, 1\}$, partial sums may be half-integers and the unit in the last place (ULP) equals $1/2$ only for magnitudes in $[512, 1024)$; once the magnitude reaches $[1024, 2048)$, the ULP increases to 1, meaning half-integers can no longer be represented exactly and are rounded to the nearest integer. Therefore, the exactness threshold for half-integer sums is 1024.}
We can adjust the binary probability for $\{0, 1\}$ based on matrix sizes to ensure a significant proportion of 
$c^{\text{ref}}_{ij}$ is below 2048 but larger than 0. 
We repeat this process multiple times with different random inputs; if any single iteration fails, the kernel is considered incorrect.

\subsection{Evaluation}
To benchmark a custom kernel against a reference kernel, 
we first define the single-run speedup $s(\text{custom})$ as:
\begin{equation}
s(\text{custom}) = \frac{t_{\text{ref}}}{t_{\text{custom}}} - 1
\end{equation}
where $t_{\text{ref}}$ and $t_{\text{custom}}$ denote the execution time of the reference and custom kernels, respectively.
Each  run  execute both the custom and reference kernels. 
To account for GPU performance variability, 
each evaluation runs for a minimum of 30 seconds
after a 10-second warmup period.
In each iteration, we randomize  execution order to eliminate ordering effect. 
The final evaluation score is the mean speed score over all runs.

\subsubsection{Offline v.s. Server Mode}
Following MLPerf \cite{mattson2020mlperf, reddi2020mlperf}, a widely-used benchmark criteria for machine learning performance, 
kernels
evaluations are performed under two scenarios: (1) {\bf offline}, where kernels are executed back-to-back without pauses, and (2) {\bf server}, where kernels are executed at intervals to simulate real-time inference. It’s worth noting that {\it the interval time is not included in the execution time used for the speedup comparison} in server mode. 
The { offline} scenario measures peak throughput by keeping the GPU fully busy, while the { server} scenario reflects real-world deployment where requests arrive at unpredictable times. Performance differs between them because back-to-back execution in { offline} mode keeps GPU caches warm with GPU running at full speed, 
while  in server mode, GPU's caches can cool down and then start from a cold stat. 

\section{CUDA-L2 for HGEMM}
\subsection{CUDA-L2}  

CUDA-L2 extends CUDA-L1 with the following key technical enhancements: 
 (1) continued pretraining with more diverse CUDA code; (2) multi-stage RL training progressing from broad (general kernel types) to specialized domains (matmul-specific kernels); (3) including more comprehensive NCU profiling metrics (e.g., memory throughput, SM occupancy, cache efficiency) in the context for better optimization decisions; and (4) incorporating retrieval-augmented context to accommodate new knowledge or architectural characteristics not covered in the foundation model.

\subsubsection{General Kernel RL}
In the continued pretraining stage, we 
collected roughly 1K CUDA kernels
 from established libraries, covering a large range of operations including linear algebra, convolution operations, reduction operations, 
element-wise operations, attentions, sampling, and others 
that cannot be readily categorized 
(e.g., embedding lookups,
 loss functions,
gradient clipping,
optimizer steps). 
Each CUDA kernel is paired with an official/successful implementation from Pytorch, ATen, CUTLASS, etc. With reference implementations of these kernels, we train a general-kernel LLM-guided RL with the reward being the 
average speedup score across all test iterations. 
CUDA-L2 adopts a
 contrastive RL training strategy
described in \cite{li2025cuda},
where the model is prompted to perform comparative analysis of previously generated CUDA variants and their execution performances.
 GRPO \cite{guo2025deepseek, shao2024deepseekmath} is adopted for LLM parameter updates. 
As suggested in \cite{li2025cuda}, rewards will be smoothed and clipped to alleviate the effect of reward hacking during RL training.

\subsubsection{HGEMM RL}
In the last stage, we continue to train RL by limiting kernels to HGEMM with varying configurations of M, N, K.
The contrastive RL strategy is adopted.

\paragraph{Reward} The reward for a generated HGEMM kernel is computed by averaging speedup scores across all test iterations, similar to the evaluation section. To foster correctness, we also penalize 
the reward by numerical deviation from the FP32 CPU ground truth.
Code length will further be penalized, pushing the model to generate concise code.

\begin{equation}
r(\text{custom}) = \frac{1}{N} \sum_{i=1}^{N} \left[\frac{t^{i}_{\text{ref}}}{t^{i}_{\text{custom}}} - \alpha \cdot \text{diff}^i \right]- \beta \text{L}(\text{custom})
\end{equation}
where $\text{diff}^i = \max_j |\text{out}_{\text{FP32}}^{i}[j] - \text{out}_{\text{custom}}^{i}[j]|$ is the maximum element-wise absolute difference and $\alpha > 0$ is the penalty coefficient.
L(\text{custom}) denotes the length of the generated code and $\beta>0$ is the associative penalty coefficient. 

NCU (NVIDIA Nsight Compute) profiling metrics are incorporated in the context for HGEMM training, enabling the model to learn from detailed performance characteristics such as memory throughput, compute utilization, warp occupancy, and cache hit rates rather than relying solely on end-to-end execution time.

Generated HGEMM kernels will be parsed into {\it .cu} files and compiled with nvcc. Therefore, CUDA C/C++, CuTe, inline PTX assembly, CUDA intrinsics, and CUTLASS templates can be used, but not Python-based DSLs like Triton.

\section{Results}
In this section, we present the performance of CUDA-L2 against baseline kernels.

\begin{table}[h]
\centering
\includegraphics[scale=0.4]{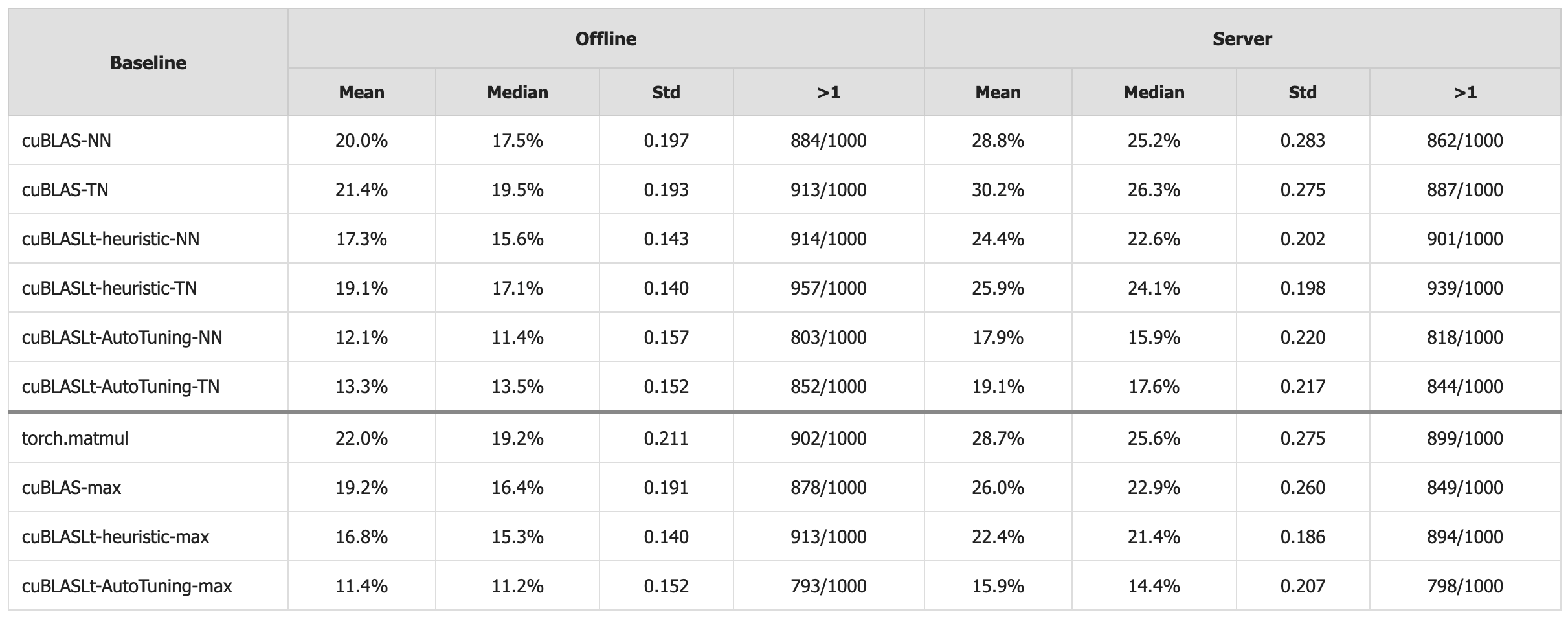}
\caption{Performance comparison of CUDA-L2 against baseline implementations (speedup ratio). 
Offline denotes that case where kernels are executed back-to-back without pauses, and server denotes kernels being executed at random intervals to simulate real-time inference.
max denotes comparing against the best kernel across TN and NN for each configuration.}
\label{major-results}
\end{table}

\subsection{Overall Results}

\paragraph{Comparisons within Baselines} Results are shown in Table \ref{major-results}. Before examining CUDA-L2's performance, we first compare the baselines against each other. Since the table shows CUDA-L2's speedup over each baseline, a higher percentage indicates a weaker baseline.
Comparing TN (transposed-normal) and NN (normal-normal) layouts, the NN layout slightly outperforms TN across all libraries. Specifically, for {\it cuBLAS}, CUDA-L2 gains 20.0\% over NN versus 21.4\% over TN. The same pattern holds for {\it cuBLASLt-heuristic} (17.3\% vs 19.1\%) and {\it cuBLASLt-AutoTuning} (12.1\% vs 13.3\%). 

Among the baseline libraries, {\it torch.matmul} is the weakest (22.0\% speedup). {\it cuBLAS-max} performs better (19.2\%) by selecting the optimal layout per configuration. {\it cuBLASLt-heuristic} improves further (16.8\%) through algorithmic selection based on problem characteristics. The strongest baseline is {\it cuBLASLt-AutoTuning} (11.4\%), which exhaustively 
test up to 100 kernel candidates per configuration.

\paragraph{CUDA-L2 Performance}

CUDA-L2 consistently outperforms all baselines, from the commonly used {\it torch.matmul} to NVIDIA's heavily-optimized {\it cuBLASLt-AutoTuning}. In offline mode, CUDA-L2 achieves average speedups of 22.0\% over {\it torch.matmul}, 19.2\% over {\it cuBLAS-max}, 16.8\% over {\it cuBLASLt-heuristic-max}, and 11.4\% over {\it cuBLASLt-AutoTuning-max}. In server mode, these gains increase to 28.7\%, 26.0\%, 22.4\%, and 15.9\% respectively. The win rates span 79.3\% to 95.7\% across all baselines, confirming that improvements are systematic rather than driven by outliers.

It is also worth noting the difference between the offline and server modes. 
We attribute this to
the difference in GPU
 thermal dynamics for GPU environment between the two modes. In the offline model, continuous execution keeps the GPU in a steady thermal state with predictable clock behavior, while in  the server mode, idle periods between requests allow the GPU to cool, causing subsequent kernel launches to experience clock boosting followed by thermal throttling. The explains the larger variance of the server mode v.s. offline mode.

\section{Optimization Techniques Discovered and Applied by CUDA-L2}
In this section, we describe CUDA optimization techniques discovered and applied by CUDA-L2.


\subsection{Abstraction Selection}
We first observe that CUDA-L2 automatically selects appropriate implementation abstractions based on $(M, N, K)$. For smaller matrices, it generates lightweight kernels directly using raw WMMA. These kernels 
 employ simpler memory access patterns, fewer pipeline stages, and minimal synchronization, which lead to faster speed.
 
For larger matrices, CUDA-L2 leans towards using {\it CuTe}'s abstractions to manage sophisticated tiled MMA operations with multi-stage pipelining. 
 CuTe's abstractions result in more concise code compared to raw CUDA intrinsics. Since the RL reward favors shorter code, CUDA-L2 naturally leans towards CuTe for complex optimizations. This allows CUDA-L2 to
 more easily
  explore advanced techniques such as software pipelining, tile shape tuning, and swizzle patterns, which yield significant speedups for large matrices.
 
 \subsection{Padding the Input Matrix with Zeros}
Matrix multiplication on GPUs is typically implemented using a tiled approach, where the two input matrices are partitioned into smaller tiles that can be efficiently processed by thread blocks. Each thread block loads a tile of the input matrices into shared memory, performs local computation, and writes the result back to global memory. 

In this tiled approach, each thread block computes a $\mathrm{BM} \times \mathrm{BN}$ output tile. A common constraint is that the matrix dimension $M$ must be divisible by $\mathrm{BM}$ and $N$ must be divisible by $\mathrm{BN}$; 
otherwise, thread blocks handling boundary tiles may issue out-of-bounds memory accesses. For example, when $M = 8192$, typical choices include $\mathrm{BM} \in \{64, 128, 256\}$.

We observe an interesting strategy discovered by CUDA-L2: setting $\mathrm{BM}$ to values where $\mathrm{M}$ is not divisible by $\mathrm{BM}$, then padding the input matrix with zeros to make $\mathrm{M}$ larger to satisfy the divisibility constraint.
This provides the flexibility to explore a bigger set of $\mathrm{BM}$ values beyond those that evenly divide the original dimension. While padding introduces computational overhead from processing additional rows, the RL model automatically finds the sweet spot of
balancing this overhead and the performance gains  by selecting a more effective $\mathrm{BM}$.

For example, in the generated kernel for (M=8192, N=512, K=2048) , CUDA-L2 selects $\mathrm{BM} = 160$, which does not divide 8192, and pads $M$ to 8320 (about 1.6\% overhead), ultimately outperforming the conventional choice of $\mathrm{BM} = 128$.
The speedup over {\it cublaslt-AutoTuning-TN} is +15.2\% when $\mathrm{BM}$ is set 160, and decreases to 0.4\% when  $\mathrm{BM}$ is set 128 and -15.7\% when $\mathrm{BM}$ is set 256. 
 
 \subsection{Proficiency in Using CUDA Optimization Techniques}
We observe that CUDA-L2 
is
proficient in using the following CUDA optimization techniques: 
\begin{itemize}[itemsep=4pt]
\item \textbf{Shared memory with bank conflict avoidance}, which uses a swizzle pattern to reorganize data layout, preventing conflicts from multiple threads accessing the same memory bank. It is parameterized by three parameters: \texttt{bits}, \texttt{base}, and \texttt{shift}, controlling the bit-level permutation of shared memory addresses.
\item {\bf Multi-stage pipelining}, which partitions shared memory into multiple buffering stages so that  during the time that
 the current tile is being computed, subsequent tiles are prefetched into other buffers. It is parameterized by the number of buffering stages, denoted by \texttt{n\_stage}. 
\item {\bf Asynchronous memory copy}, which enables non-blocking 128-bit transfers from global to shared memory.
\item {\bf Register accumulation}, which stores partial results in register files to minimize memory traffic.
\item {\bf Block swizzle}, 
which improves L2 cache hit rates by reordering thread block execution so that spatially adjacent blocks can run concurrently, 
 while by default, thread blocks are launched in row-major order, causing adjacent blocks to access disjoint memory regions resulting in poor L2 cache utilization.
 Block swizzle  is parameterized by {\it swizzle\_stride}, which determines the stride pattern used to reorder block indices.
\item {\bf Epilogue optimization}, which efficiently transfers accumulated results from registers to global memory, potentially using direct register-to-shared-memory copies with wide data types (e.g., 128-bit) to minimize intermediate temporaries and reduce memory traffic.
\end{itemize}

With these techniques, CUDA-L2 can also automatically determine which to apply and figure out their optimal parameters. For instance, it decides whether to use block swizzle or bypass it entirely, whether to enable software pipelining or rely on simple loops, and how to set tile sizes, pipeline stages, etc.

Moreover, within each technique, CUDA-L2 can discover novel variations beyond standard implementations. Rather than applying a one-size-fits-all strategy, these variations target specific MNK configurations where they outperform conventional approaches. 
The following are a few examples:

\section{Analysis}
In this section, we conduct a systematic analysis of the 1,000 optimal configurations discovered by RL to 
understand which optimization techniques and parameter settings should be applied under different problem dimensions and hardware constraints.
\begin{table}[h]
\centering
\includegraphics[scale=0.4]{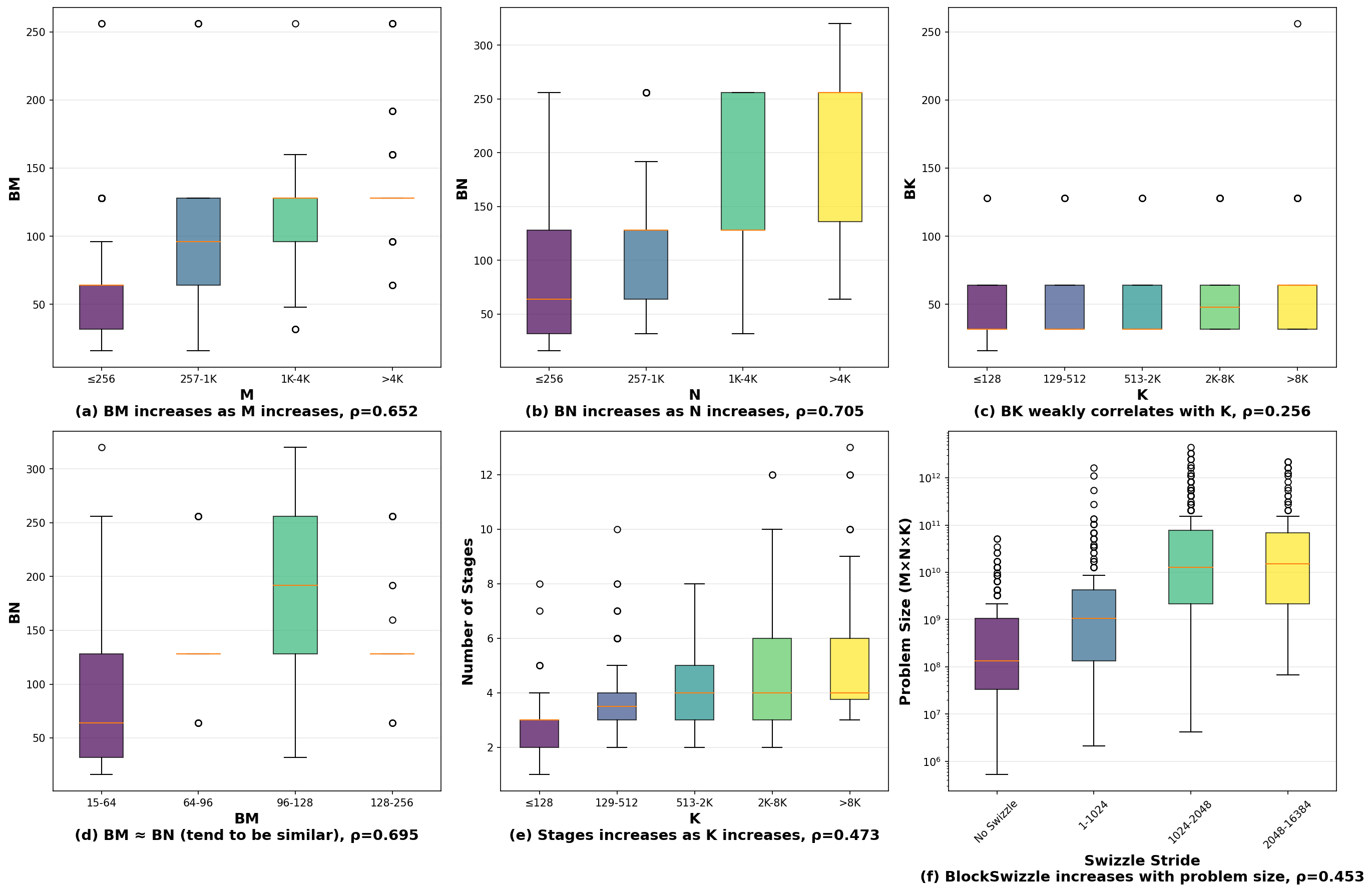}
\caption{Hyperparameter Selection Patterns in Optimized CUDA Matrix Multiplication Kernels}
\label{fig:stage}
\end{table}
\subsection{How to Choose BM, BN, BK}
Figure \ref{fig:stage}(a)(b)(c) illustrates the relationship between problem dimensions and their corresponding tile sizes parameterized by BM, BN and BK, which determine how the input matrices are divided into smaller blocks for efficient computation on the GPU.

Figure \ref{fig:stage}(a) shows that BM scales proportionally with the M dimension ($\rho=0.652$). As M increases from small values ($\leq$256) to large values ($>$4K), BM grows from approximately 60 to 160. This strong positive correlation indicates that larger M dimensions require larger M-tiles to maintain efficiency.
A similar trend is observed between N and BN in Figure \ref{fig:stage}(b), with an even stronger correlation ($\rho=0.705$). 
In contrast, Figure \ref{fig:stage}(c) reveals that BK has only a weak correlation with K ($\rho=0.256$), indicating that BK selection depends on additional factors beyond K alone, such as register pressure, memory bandwidth constraints, and the number of pipeline stages.

Figure \ref{fig:stage}(d) reveals that the values of BM and BN are highly correlated ($\rho=0.695$), indicating that optimal configurations tend to use similar or same values for both tile dimensions. 
This is because the tensor cores operate in square or near-square instruction format (e.g., 16$\times$8$\times$16), and that 
balanced tiles reduce the risk of resource imbalance where one dimension becomes a bottleneck. 

\subsection{How to Choose Stage Number in Multi-stage Pipelining}
Figure \ref{fig:stage}(e) shows how matrix dimensions 
affect the number of pipeline stages in optimized CUDA kernels. Each box shows the distribution of dimension values for different stage counts.
It shows a clear positive relationship: as the K dimension increases, the number of pipeline stages should increase to achieve the optimal performance. 
Small K ($\leq$128) needs only 2-3 stages for adequate latency hiding, while large K ($>$8K) requires 6+ stages to maintain high throughput by keeping multiple data loads in flight simultaneously.

\subsection{When and How to Use Block Swizzling}
BlockSwizzle is a memory optimization technique that improves cache locality and load balancing by reordering how thread blocks access global memory. 
Figure \ref{fig:stage}(f) shows the relationship between matrix problem ($M\times N\times K$) and BlockSwizzle usage across different  configurations. The first box represents configurations that do not use BlockSwizzle, which typically handle smaller problem sizes. The subsequent boxes show increasingly larger swizzle stride values, with each handling progressively larger problems.

The decision of when to enable BlockSwizzle is primarily driven by problem size. For small problems (less than $2^{27}$ or approximately 134 million operations), BlockSwizzle is optional and used in only 44\% of configurations, as the overhead may outweigh the benefits when data fits comfortably in cache. However, as problem size grows, BlockSwizzle becomes increasingly prevalent: medium problems ($2^{27}$ to $2^{33}$) use it 73-80\% of the time, while very large problems (greater than $2^{36}$ or 68 billion operations) employ it almost every time at 99\% usage.

The value of {\it swizzle\_stride} parameter, which controls the granularity of the block reordering, should be chosen adaptively based on problem dimensions. Smaller problems that do use swizzling typically employ stride values of 8-128, while larger problems benefit from stride values of 512-16,384. The strong correlation ($\rho=0.453$) demonstrates that both the decision to use BlockSwizzle and the selection of an appropriate stride value are fundamentally tied to problem scale, with the optimization becoming essential as memory access patterns grow more complex at larger scales.

\section{Conclusion}

In this paper, we propose CUDA-L2, a system that combines large language models and reinforcement learning to automatically optimize HGEMM CUDA kernels across diverse configurations. Through continued pretraining on diverse CUDA code, multi-stage RL training progressing from general kernel optimization to matmul-specific optimization, and retrieval-augmented context, CUDA-L2 achieves an 11.4\% speedup over NVIDIA's {\it cuBLASLt-AutoTuning} library on A100 GPUs across 1000 configurations in offline mode, and 15.9\% speedup in server mode. Against more commonly used baselines, CUDA-L2 delivers even larger gains: 22.0\% over {\it torch.matmul} (28.7\% server) and 19.2\% over {\it cuBLAS} (26.0\% server).
CUDA-L2 demonstrates that even for heavily-optimized, performance-critical kernels like HGEMM, LLM-guided RL can discover superior implementations by systematically exploring configuration spaces at scales impractical for manual optimization. 

\bibliography{custom}
\bibliographystyle{acm}
\newpage

\end{document}